\documentclass[journal]{IEEEtran}
\usepackage{tabularray}
\usepackage{siunitx}
\usepackage{enumerate}
\makeatletter
\def\endthebibliography{%
	\def\@noitemerr{\@latex@warning{Empty `thebibliography' environment}}%
	\endlist
}
\makeatother

\usepackage[noadjust]{cite}

\ifCLASSINFOpdf
  \usepackage[pdftex]{graphicx}
  \graphicspath{{pdf/}{jpg/}{png/}}
\else
\fi
\usepackage{amsmath}
\usepackage{cases}

\usepackage{array}
\usepackage{caption}
\usepackage{float}
\usepackage{tabularray}
\usepackage{algorithm}
\usepackage{algpseudocode}
\usepackage{booktabs}
\usepackage{stfloats}
\usepackage{multirow}
\usepackage{tabularray}
\fnbelowfloat
\usepackage{xspace} 
\AtBeginDocument{%
	\let\OLDdegree\degree%
	\renewcommand{\degree}{\OLDdegree{}\xspace}%
}

\usepackage{xcolor} 

\usepackage{footnote}
\makesavenoteenv{tabular}
\makesavenoteenv{table}

\newif\ifLPRfmtDRAFT
\newif\ifLPRfmtHIGHLIGHTS
\newif\ifLPRfmtNOAUTH

\newif\ifNO
\newif\ifYES
\YEStrue

\LPRfmtHIGHLIGHTStrue
\LPRfmtDRAFTtrue

\ifLPRfmtDRAFT

	\usepackage[colorinlistoftodos]{todonotes}
	\usepackage{mdframed}
	\usepackage{scrtime}
	\usepackage{soul}

	\mdfdefinestyle{notesstyle}{%
		linecolor=red,frametitlerule=true,
		backgroundcolor=red!15}
	\mdfdefinestyle{outlinestyle}{%
		linecolor=green,frametitlerule=true,
		backgroundcolor=green!15}
	
	\presetkeys{todonotes}{color=blue!15, linecolor=red!90, size=\footnotesize, figwidth=3.4in}{}
	\makeatletter
	\if@todonotes@disabled
		\newcommand{\hlfix}[2]{#1}
	\else
		\newcommand{\hlfix}[2]{\texthl{#1}\todo{#2}{}}
	\fi
	\makeatother
\else
	\makeatletter
	\newcommand{\hlfix}[2]{{}}
	\newcommand{\todo}[1]{{}}
	\makeatother
\fi

\ifLPRfmtHIGHLIGHTS
	\colorlet{editColorBlue}{blue!100!}
	\colorlet{editColorRed}{red!100!}
	\definecolor{editColorGreen}{RGB}{0,127,0}
	\definecolor{editColorRedRed}{RGB}{192,0,128}
	\definecolor{editColorBlueBlue}{RGB}{0,96,192}

\else

\fi

\begin{document}

\title{A Driving Regime-Embedded Deep Learning Framework for Modeling Intra-Driver Heterogeneity in Multi-Scale Car-Following Dynamics}

\ifLPRfmtNOAUTH
\author{}
\else
\author{
        Shirui Zhou,
	Jiying Yan,
	Junfang Tian*,  
        Tao Wang, 
        Yongfu Li, \IEEEmembership{Senior~Member,~IEEE,}
        Shiquan Zhong
        \thanks{This work is supported by the National Natural Science Foundation of China (Grant No. 72222021, W2411064, 72431006, 72288101).}
        \thanks{Shirui Zhou, Jiying Yan, Junfang Tian and Shiquan Zhong are with Institute of Systems Engineering, College of Management and Economics, Tianjin University, Tianjin, China.(e-mail: jftian@tju.edu.cn).Tao Wang is with College of Automation and Electronic Engineering, Qingdao University of Science and Technology, Qingdao, China. Yongfu Li is with College of Automation, Chongqing University of Posts and Telecommunications, Chongqing, China.}
        }


\markboth{}{Zhou \MakeLowercase{\textit{et al.}}: \<Hybrid Car-Following Modeling with Driving Regimes Embedding\>}

\maketitle

\begin{abstract}
A fundamental challenge in car-following modeling lies in accurately representing the multi-scale complexity of driving behaviors, particularly the intra-driver heterogeneity where a single driver's actions fluctuate dynamically under varying conditions. While existing models, both conventional and data-driven, address behavioral heterogeneity to some extent, they often emphasize inter-driver heterogeneity or rely on simplified assumptions, limiting their ability to capture the dynamic heterogeneity of a single driver under different driving conditions. To address this gap, we propose a novel data-driven car-following framework that systematically embeds discrete driving regimes (e.g., steady-state following, acceleration, cruising) into vehicular motion predictions. Leveraging high-resolution traffic trajectory datasets, the proposed hybrid deep learning architecture combines Gated Recurrent Units for discrete driving regime classification with Long Short-Term Memory networks for continuous kinematic prediction, unifying discrete decision-making processes and continuous vehicular dynamics to comprehensively represent inter- and intra-driver heterogeneity. Driving regimes are identified using a bottom-up segmentation algorithm and Dynamic Time Warping, ensuring robust characterization of behavioral states across diverse traffic scenarios. Comparative analyses demonstrate that the framework significantly reduces prediction errors for acceleration (maximum MSE improvement reached 58.47\%), speed, and spacing metrics while reproducing critical traffic phenomena, such as stop-and-go wave propagation and oscillatory dynamics. 
\end{abstract}%


\IEEEpeerreviewmaketitle

\section{Introduction}

Understanding and modeling car-following behavior is fundamental to microscopic traffic flow research, serving as a bridge between individual driving dynamics and macro-level traffic phenomena. Traditional physics-based car-following models, such as the Optimal Velocity (OV) model \cite{bando_dynamical_1995}, Intelligent Driver Model (IDM) \cite{treiber_congested_2000}, Full Velocity Difference Model (FVDM) \cite{jiang_full_2001}, and Newell model \cite{newell_simplied_2002}, rely on predefined mathematical formulations with interpretable parameters to describe vehicle longitudinal dynamics based on spacing, time headway, speed, and speed differentials. While these models offer theoretical insights and computational efficiency, their fixed structures and parameter constraints make them hard to capture the variability and complexity of real-world driving behaviors. This limitation highlights the need for more adaptive frameworks capable of accommodating diverse and dynamic driving patterns.

Recent advances in data collection and processing have facilitated the development of data-driven approaches that leverage machine learning and deep learning techniques to learn interaction patterns directly from large-scale trajectory datasets, which can achieve impressive results \cite{liTrajectoryDatabasedTrafficFlow2020}. Conventional machine learning models, such as Support Vector Regression \cite{wei_analysis_2013} and Locally Weighted Regression \cite{papathanasopoulou_towards_2015}, approximate nonlinear relationships, while deep learning methods, including Artificial Neural Networks \cite{zheng_car-following_2013} and Deep Neural Networks \cite{huang_car-following_2018}, excel in capturing temporal dependencies and high-dimensional interactions. For instance, \cite{wu_memory_2019} proposed a deep learning framework incorporating memory retention, attention mechanisms, and anticipation modules for car-following modeling. Reinforcement learning frameworks \cite{zhu_human-like_2018} further enhance behavioral modeling by mimicking human decision-making. Among these, deep learning-based architectures have demonstrated exceptional efficacy in capturing nonlinear, context-dependent dynamics, marking a paradigm shift from traditional physics-based models to more adaptive and flexible frameworks \cite{ma_sequence_2020,yuImpactPriorExperiencesCarFollowing2023}. 

Despite these advancements, a key challenge remains: addressing the \textit{behavioral heterogeneity} inherent in driving behaviors. In \cite{wangCapturingCarFollowingBehaviorsDeep2018}, a data-driven paradigm was proposed to reduce manual intervention by relying entirely on deep learning models. The study demonstrated that sufficiently large LSTMs or GRUs can model car-following dynamics if high-quality and abundant data are available. However, it briefly emphasized the need to address behavioral heterogeneity to distinguish individual driving characteristics. Behavioral heterogeneity, defined as the variability in drivers' responses to similar stimuli, is broadly categorized into \textit{inter-driver heterogeneity} (differences between drivers) and \textit{intra-driver heterogeneity} (the internal stochasticity of an individual driver). Both forms of heterogeneity significantly influence traffic dynamics, including response latency, time headway preferences, and traffic oscillations \cite{saifuzzaman_incorporating_2014,laval_hysteresis_2011,chen_periodicity_2014,chen_behavioral_2012,chen_microscopic_2012}.

Inter-driver heterogeneity focuses on differences between drivers' behavioral characteristics. For example, \cite{ossen_heterogeneity_2011} distinguished between "driving style heterogeneity" (behavioral rules differing across drivers) and "within-style heterogeneity" (parameter variations under the same rule), emphasizing the nuanced drivers of behavior. \cite{makridisFormalizingHeterogeneityVehicledriverSystem2020} further demonstrated that explicitly modeling driver heterogeneity—including variations in driving styles and vehicle dynamics—can replicate traffic oscillations and instabilities without artificial noise injection. Similarly, studies such as \cite{chen_investigating_2020} incorporated inter-driver variability into traffic flow models via parameter-space clustering, highlighting its critical role in reproducing macroscopic phenomena. These approaches effectively address \textit{inter-driver variability}, particularly at the population level.

Intra-driver heterogeneity, by contrast, represents variability within a single driver under different conditions and is often dynamic and context-dependent. For instance, \cite{ossen_multi-anticipation_2006} revealed significant intra-driver variability, where individual drivers exhibit parameter fluctuations across different scenarios, reflecting the stochasticity of human decision-making. \cite{wagner_analyzing_2012} observed that fluctuations in car-following behavior arise primarily from inconsistent responses within individual drivers rather than differences across drivers. This form of heterogeneity is critical to understanding traffic phenomena like stop-and-go waves and traffic oscillations. For example, \cite{laval_mechanism_2010} established a theoretical framework explaining how interactions between aggressive and conservative driving behaviors generate traffic instabilities, while \cite{huangExperimentalStudyModelingCarfollowing2018} demonstrated that accounting for intra-driver heterogeneity is essential for replicating these dynamics. Drivers exhibit distinct car-following patterns under varying conditions, such as acceleration, deceleration, and cruising. For instance, \cite{wang_driving_2019} observed significant behavioral differences in 65\% of drivers between acceleration and deceleration phases, while \cite{chen_behavioral_2012} revealed shifts in spatiotemporal lag parameters across congested and non-congested conditions. Dynamic Time Warping analyses by \cite{taylor_method_2015} further highlighted time-varying parameter changes under different traffic conditions. Despite these insights, achieving a comprehensive framework that fully integrates intra-driver heterogeneity into predictive models remains a significant challenge.

Driving regimes, such as steady-state following, acceleration, and cruising, provide a structured framework for modeling \textit{intra-driver heterogeneity}. As highlighted by \cite{treiberTrafficFlowDynamics2013}, a complete driving trajectory can be segmented into six distinct regimes: free acceleration, cruising at the desired speed, following a leader at constant speed, accelerating behind a leader, decelerating behind a leader, and standing behind a leader. This segmentation enables detailed modeling of temporal transitions in driver behavior. For example, \cite{sharma_is_2019,sharma_pattern_2018,taylor_method_2015} classified driving modes in the NGSIM dataset into car-following and free-flow states, reflecting the stimulus-response mechanisms underlying car-following behavior. Building on this, \cite{ma_behavior_2007} proposed a data-driven framework that employs a fuzzy clustering algorithm with temporal continuity constraints, significantly improving the classification accuracy of driving regimes (e.g., acceleration, braking) in complex scenarios like lane-merging and vehicle separation. Additionally, \cite{ma_statistical_2007} conducted a statistical analysis of car-following behavior across different regimes, demonstrating that hybrid modeling approaches—combining linear models for straightforward regimes (e.g., braking) with nonlinear methods for oscillatory regimes (e.g., stable following)—can markedly enhance behavioral realism. Complementarily, \cite{xuStatisticalInferenceTworegimeStochastic2020} extended the Newell model by introducing a two-state (free-flow and congested-flow) stochastic car-following model, where independent stochastic processes were used to characterize variability in both free-flow and congestion dynamics, effectively capturing intra-driver heterogeneity. Despite these advancements, most car-following models—whether classical or data-driven—seldom incorporate driving regimes directly into their predictive frameworks. Instead, many studies rely on post hoc clustering or focus on \textit{inter-driver heterogeneity} at the population level \cite{ma_sequence_2020,zhou_recurrent_2017,lu_learning_2023,xu_sequence--sequence_2024}. While some recent efforts, such as \cite{ma_statistical_2007} and \cite{xuStatisticalInferenceTworegimeStochastic2020}, have introduced multi-regime frameworks or stochastic models to address the role of driving regimes, these approaches remain limited in scope and are not yet widely adopted. This oversight of \textit{intra-driver variability} reduces the adaptability of existing models to diverse traffic scenarios and limits their utility in control-oriented applications. Explicitly embedding driving regimes into predictive frameworks represents a promising avenue to better capture the dynamic and context-dependent nature of individual driving behavior, thereby enhancing the accuracy of traffic flow predictions.

The contributions of this study are four-fold. First, a novel data-driven car-following framework is proposed, which integrates driving regimes into the modeling process to address intra-driver heterogeneity. A bottom-up segmentation method is introduced to classify trajectory sequences into discrete driving regimes based on kinematic parameters, with the Dynamic Time Warping (DTW) algorithm applied to refine regime classification for robust handling of dynamic driving behaviors under varied traffic conditions. Second, a hybrid deep learning architecture is developed, combining Gated Recurrent Units (GRU) to capture discrete regime transitions and LSTM networks to model continuous motion dynamics. This architecture seamlessly integrates discrete decision-making processes with continuous vehicle dynamics, providing a comprehensive representation of intra-driver variability and nonlinear car-following interactions. Third, the proposed framework outperforms classical physics-based models (e.g., IDM) and deep learning approaches, achieving significant improvements in prediction accuracy and computational efficiency. Finally, it successfully reproduces critical traffic phenomena, such as stop-and-go wave propagation and oscillatory dynamics, validating its ability to capture both microscopic driving behaviors and macroscopic traffic patterns.

The remainder of this article is organized as follows. Section II introduces the development of the proposed framework. It begins with a description of the dataset, followed by the classification of driving regimes, which includes bottom-up segmentation for speed profile characterization, identification of car-following and free-flow sections, and operational submode classification. This section also presents the development of the DR-embedded data-driven model, covering its architecture, configuration, and training. Section III focuses on model evaluation and comparison, including micro-level characteristic analysis to assess driver-specific behaviors and macro-level validation to evaluate traffic phenomena reproduction. Finally, Section IV summarizes the findings and suggests future research directions.

\section{Model Development}

\subsection{Data Description}

The NGSIM dataset, developed by the Federal Highway Administration, provides high-resolution (10Hz) vehicle trajectory data, making it a benchmark resource for microscopic traffic dynamics analysis. This study utilizes the I-80 freeway dataset, enhanced through Montanino and Punzo's physics-informed validation framework \cite{punzo_assessment_2011, montanino_trajectory_2015}, which rectifies issues such as velocity discontinuities and acceleration outliers. Focusing on a 400-meter homogeneous highway segment, the dataset includes 15 minutes of peak-hour trajectory data (16:00-16:15 UTC-8). For model development , 856 vehicle pairs (231,458 samples) from lanes 2-4 were used for training and validation(80\% for training and 20\% for validation) , while lane 5 data (185 pairs, 85,749 samples) was reserved for independent testing.

\subsection{Driving Regimes Classification}

Traffic flow interactions are broadly categorized into two primary driving regimes: car-following (CF) and free-flow (FF) \cite{sharma_is_2019,sharma_pattern_2018}. The CF regime involves distance-dependent interactions, where longitudinal dynamics are driven by time-delayed acceleration responses to leading vehicles. In contrast, the FF regime reflects minimal influence from leading vehicles, with behaviors primarily shaped by ambient traffic conditions. The CF regime is further divided into four phases: steady-state following (F), acceleration (A), deceleration (D), and stationary state (S), while the FF regime includes free acceleration (Fa) and cruise control (C). This classification is validated through microscopic behavior clustering, offering a theoretical basis for modeling driving state transitions.

A hybrid framework integrating bottom-up hierarchical clustering and DTW is proposed to address two challenges inherent to vehicle trajectory data: (1) temporal misalignment in asynchronous leader-follower sequences, resolved through DTW-based warping, and (2) time-lagged dependency extraction in car-following dynamics. This approach supports driving regime identification and is structured into three implementation phases:

\begin{enumerate}[i.]
    \item \textbf{Bottom-Up Spatiotemporal Feature Extraction:} Leader-follower trajectory pairs are partitioned into spatiotemporal primitives using speed profile alignment, isolating atomic behavioral units (e.g., acceleration/deceleration events).
    \item \textbf{Car-Following and Free-Flow section Identification}:A probabilistic classification framework distinguishes CF and FF regimes by analyzing feature distributions (e.g., Time delay) within trajectory primitives.
    \item \textbf{Operational Submode Identification:} Acceleration/deceleration submodes (A/D) and cruise states (C) are classified using statistically derived kinematic thresholds (e.g., $\pm0.5\text{m/s}^2$ for steady-state following) and steady-state speed maintenance criteria.
\end{enumerate}

\subsubsection{Bottom-Up Spatiotemporal Feature Extraction}

To extract the spatiotemporal feature of trajectories, we introduce a trajectory segmentation framework for transforming continuous speed profiles into interpretable kinematic states. Grounded in the hierarchical time-series segmentation principles of \cite{last_segmenting_2004}, the algorithm iteratively merges temporal segments by optimizing kinematic homogeneity (e.g., acceleration/deceleration consistency), validated through residual error minimization. The implementation steps are as follows:

\begin{enumerate}[i.]
    \item \textbf{Initialization:} Discretize the raw speed time-series data  $V=\left\{v\left(t_i\right) | i=\right.$ $1,2,..., N\}$ into atomic segments  $S=\left\{s_k | s_k=[t_k, t_{k+1}], k=1,..., N-1\right\}$
    \item \textbf{Cost Computation:} Calculate merging costs for adjacent segment pairs using a dual-criterion cost function: 
    \begin{equation}
        \operatorname{cost}\left(s_j, s_{j+1}\right)=\left\|\Delta \theta_\text{L2}\right\|-\lambda \operatorname{sgn}\left(\theta_i \cdot \theta_j\right)
    \end{equation}
    where $\theta_i$ and $\theta_j$ denote linear regression slopes of respective segments, $\left\|\Delta \theta_\text{L2}\right\|$ represents the Euclidean distance between slopes  and $\lambda$ penalizes sign-alternating slope pairs (e.g., acceleration-to-deceleration transitions) to detect abrupt driving behavior. sgn represents the signum function.
    \item \textbf{Iterative Merging:} Merge the lowest-cost segment pairs iteratively until meeting the termination criterion:
    \begin{equation}
    N_{\max }=\left\lfloor\frac{T}{L_{\min }}\right\rfloor=\lfloor 2 T\rfloor
    \end{equation}
    where $T$ is the total observation duration (in seconds), and  $L_{\text {min }}=0.5 \mathrm{~s}$    ensures empirically valid segment lengths aligned with the NGSIM dataset's 10 Hz sampling rate and prior methodologies \cite{sharma_is_2019,sharma_pattern_2018}.
\end{enumerate}

To address inherent limitations in state persistence identification (e.g., fragmented acceleration/deceleration phases due to minor slope fluctuations), we propose a post-processing refinement step: 

\textbf{State interval merging}: Consecutive segments  $s_j, s_k \in S^{\prime}$ ith slope differences below a tolerance threshold  $\epsilon=0.01s$ are merged into kinematically coherent intervals: 
\begin{equation}
   \begin{aligned}
\forall s_j, s_k & \in S^{\prime}, i f\left|\theta_i-\theta_k\right|<\varepsilon \\
S^{\prime} & =S^{\prime}\left\{s_j, s_k\right\} \cup\left\{\operatorname{merge}\left(s_j, s_k\right)\right\}
\end{aligned} 
\end{equation}

where $\theta_j$ and $\theta_k$ denote the linear regression slopes of segments $s_j$ and $s_k$, respectively. This optimization consolidates transient kinematic variations (e.g., noise-induced slope deviations) while preserving physically interpretable driving state durations, validated against NGSIM ground-truth annotations \cite{taylor_method_2015}.

\subsubsection{Car-Following and Free-Flow Section Identification}

To classify the CF and FF sections, we propose a framework combining Newell's car following theory \cite{newell_simplied_2002,xuStatisticalInferenceTworegimeStochastic2020} with DTW using segmented speed profiles. The framework establishes leader-follower interaction dynamics through two complementary mechanisms:

\begin{itemize}
    \item \textbf{Congested Section}: Newell's model defines follower kinematics as time- and space-displaced leader trajectories:
    \begin{equation}
            x_n\left(t+\tau_n\right)=\min \{{x_n(t)+\underbrace{V_0 \times \tau_n}_{\text {free-flow}}, \underbrace{x_{n-1}(t)-d_n}_{\text{congestion}}}\}
    \end{equation}
    where $\left(\tau_n\right)$ is the time delay, $\left(d_n\right)$ denotes the minimum safety spacing, and $\left(V_0\right)$ is the desired free-flow speed.  Time delay $\tau_n$ is aggregates driver reaction time $\tau_{\text{react}}$ and vehicle mechanical response delay $\tau_{\text{mech}}$, reflecting total latency in car-following adjustments,and minimum safety spacing $d_n$: Minimum distance spacing between vehicles, determined by kinematic constraints and driver risk perception.
    \item \textbf{Free-Flow Section}: Vehicles maintain  desired speed $V_0$ with negligible leader influence, modeled as:
\begin{equation}
    \frac{dx_n(t)}{dt}=V_0
\end{equation}
\end{itemize}

Regime transitions are identified through discontinuities in $\tau_n$ and $d_n$, establishing the theoretical basis for differentiating CF and FF sections. To operationalize this framework, we apply DTW to align leader-follower trajectories, resolving nonlinear time shifts and quantifying stimulus-response patterns as validated in transportation behavior studies \cite{sharma_is_2019,sharma_pattern_2018}. The workflow consists of three stages:

\textbf{1. Data Preprocessing}

Synchronize leader (L) and follower (F) trajectory pairs to construct temporally aligned sequences of speed $v_\text{L}(t), ~ v_\text{F}(t)$ and position $x_\text{L}(t), ~ x_\text{F}(t)$.

\textbf{2.Dual-Modal DTW Alignment}
\begin{itemize}
    \item Velocity alignment: Compute optimal warping path $\pi$ minimizing cumulative speed discrepancy
    \begin{equation}
    \text{DTW}_v=\arg \min _\pi \sum_{(i, j) \in \pi}\left|v_L\left(t_i\right)-v_F\left(t_j\right)\right|
    \end{equation}
    \item Position alignment: Derive spatial correspondence through trajectory matching 
        \begin{equation}
   \text{DTW}_x=\arg \min _\pi \sum_{(i, j) \in \pi}\left|x_L\left(t_i\right)-x\left(t_j\right)\right|
        \end{equation}
\end{itemize}

\textbf{3. Parameter Extraction}

For each trajectory pair $k$, extract time delay $\tau_n^k$ and safety spacing $d_n^k$ from optimal warping paths:
\begin{equation}
\left\{
\begin{aligned}
\tau_n^k &=t_j^{(k)}-t_i^{(k)} \\
d_n^k &=x_\text{L}\left(t_j^{(k)}\right)-x_\text{F}\left(t_i^{(k)}\right) 
 \end{aligned}
 \right.
\end{equation}

\textbf{4. Regime Classification}

Since free-flow behavior is uncommon in typical traffic scenarios, the 85$^\text{th}$ percentile is adopted as a statistical threshold to distinguish extreme values, a method widely used in engineering applications \cite{houStatisticalTest85th15th2012}. Using this approach, CF and FF regimes are classified based on the 85$^\text{th}$ percentile threshold ($P_{\tau_n}^{85}$) of the time delay distribution $\tau_n^k$, which is empirically calibrated from the NGSIM trajectory dataset.

\begin{equation}
  \text{CF}=\left\{k \mid \tau_n^k \leq P_{\tau_n}^{85}\right\} \quad \text{FF}=\left\{k \mid \tau_n^k>P_{\tau_n}^{85}\right\}  
\end{equation}

This threshold aligns with the intrinsic behavioral dichotomy between driver-vehicle interactions in car-following and free flow traffic states. In CF section($\tau_n^k \leq P_{\tau_n}^{85}$) vehicles exhibit spacing-dependent interactions where drivers actively adjust accelerations to maintain safe headways, consistent with the stimulus-response principles of follow-the-leader dynamics. Conversely, FF section($\tau_n^k > P_{\tau_n}^{85}$) reflect driver independence from immediate leaders, characterized by stochastic speed adjustments governed by ambient traffic conditions or driver preferences rather than direct leader-follower coupling. 

\subsubsection{Operational Submode Classification}

Building on the CF and FF section segmentation, a hierarchical slope-based classifier identifies kinematic submodes within each regime.  CF section includes acceleration,deceleration,following the leader at constant speed, standing behind the leader. FF section includes free acceleration and cruising at desired speed. The framework (Fig.\ref{Fig_Slope}) comprises four stages.

\begin{figure}[htbp]
    \centering
    \includegraphics[width=0.8\linewidth]{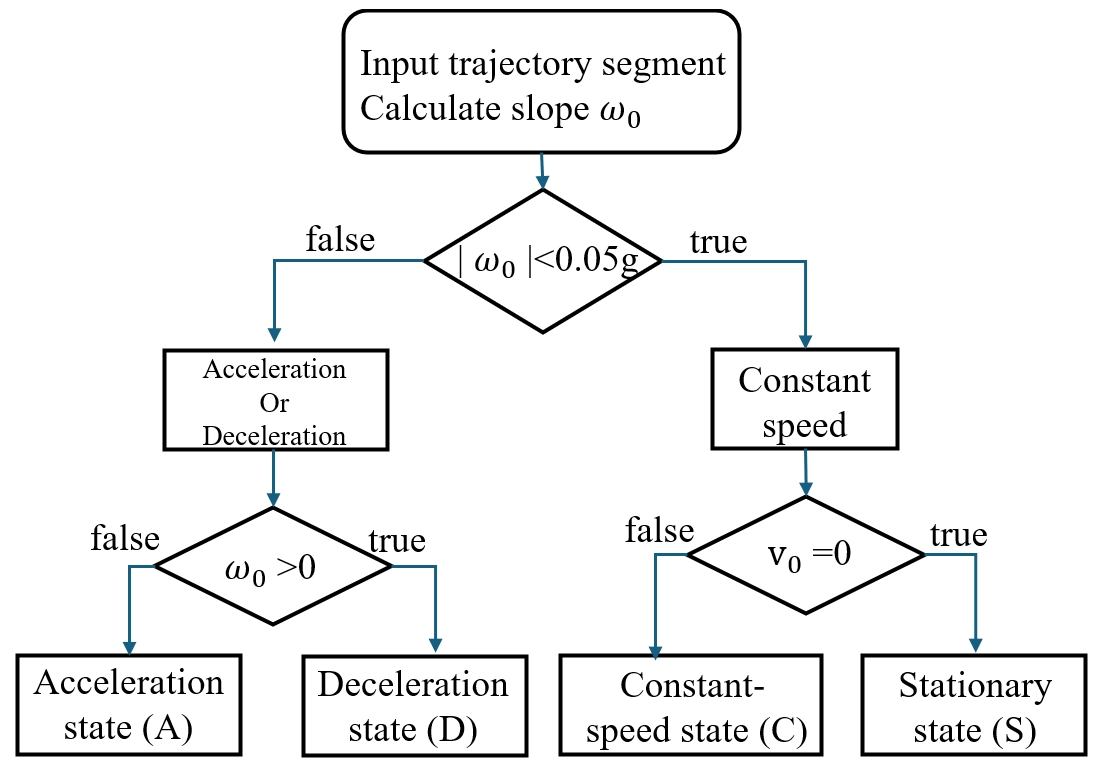}  
    \caption{Framework of Slope-based Analysis Methodology.}
    \label{Fig_Slope}
\end{figure}

The state classification framework initiates by processing segmented car-following (CF) and free-flow (FF) speed profiles with minimum 3-second duration. Slope calculation (denoted as $\omega_0$) is performed using piecewise linear regression, where the absolute slope threshold $|\omega_0|$ = 0.5 corresponds to Ozaki's stability criterion of 0.05g (0.49m/s$^2$) [46]. Segments with $|\omega_0|$ $\leq$ 0.5 are designated as steady-state conditions, requiring secondary validation through instantaneous speed $(v_0)$ verification: trajectories maintaining $v_0\neq0$ are classified as constant-speed states (C), while zero-velocity segments ($v_0=0$) are categorized as stationary states (S).

For unsteady regimes $(|\omega_0|>0.5)$, directional slope analysis bifurcates the classification: positive slopes $(\omega_0>0.5)$ indicate acceleration states (A), whereas negative slopes $(\omega_0<-0.5)$ correspond to deceleration states (D). In CF scenarios, this manifests as acceleration-following (A), deceleration-following (D) states, constant-following (F) and stationary
state(S). FF regimes exhibit similar acceleration patterns (free-acceleration, Fa) and cruising states (C), though fundamentally exclude deceleration/stationary conditions due to their traffic flow dynamics. The dual-layer validation mechanism - combining slope-driven primary classification and velocity-based steady-state confirmation - ensures kinematic state identification aligns with both data-driven patterns and theoretical stability criteria.
\begin{figure}[htbp]
    \centering
    \includegraphics[width=0.8\linewidth]{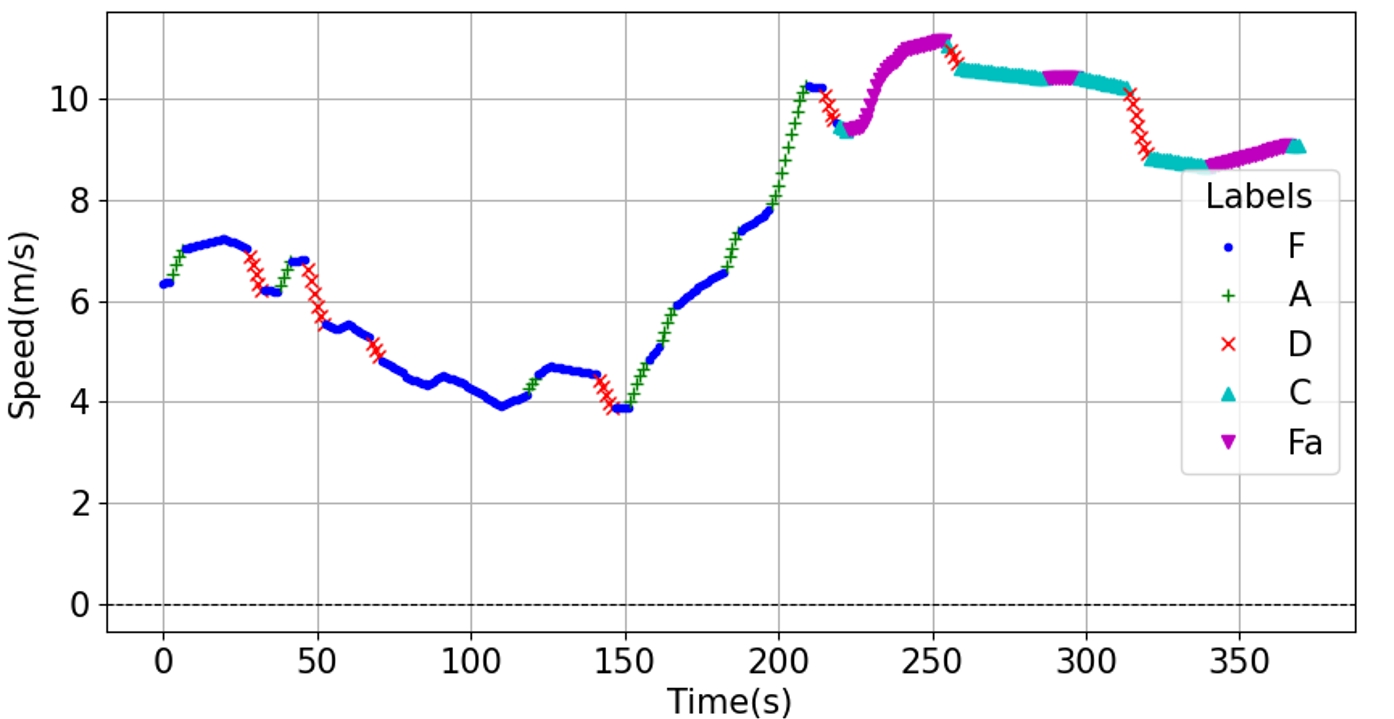}  
    \caption{One Example of Driving Regimes Identification.}
    \label{Fig_DRRS}
\end{figure}

Fig.\ref{Fig_DRRS} demonstrates empirical validation through vehicle-specific case studies. The slope thresholds achieve optimal precision-recall balance through receiver operating characteristic analysis of NGSIM validation data. 

\subsection{Development of DR-embedded Data-Driven Model}

Recent advancements in deep learning, particularly neural-based architectures, have proven effective in capturing temporal dependencies and modeling stochastic car-following behaviors. For instance, \cite{zhou_recurrent_2017} demonstrated that Recurrent Neural Networks (RNN) outperform Feedforward Neural Networks (FNN) in predicting traffic oscillations, while \cite{huang_car-following_2018} highlighted the ability of LSTM networks to replicate oscillation propagation patterns. Moreover, \cite{wang_long_2019} emphasized the strengths of GRU architectures in improving prediction consistency and addressing parameter calibration challenges. Building on these insights, this study explores the complementary strengths of GRU and LSTM structures in adapting to the discrete-continuous nature of driving dynamics. Specifically, GRU networks are well-suited for capturing transitions between discrete driving regimes (e.g., acceleration, braking), while LSTM networks excel at handling the temporal continuity of continuous features such as speed and acceleration.

\subsubsection{Model Architecture}

The proposed framework, illustrated in Fig.\ref{Fig_ModelArch}, consists of two core modules: a driving regime predictor and a kinematic feature fusion module. By iteratively predicting discrete driving states and continuous motion parameters, the architecture enables refined real-time modeling of vehicle dynamics. Technical implementation details are elaborated below.

\begin{figure*}[htbp]
    \centering
    \includegraphics[width=0.7\linewidth]{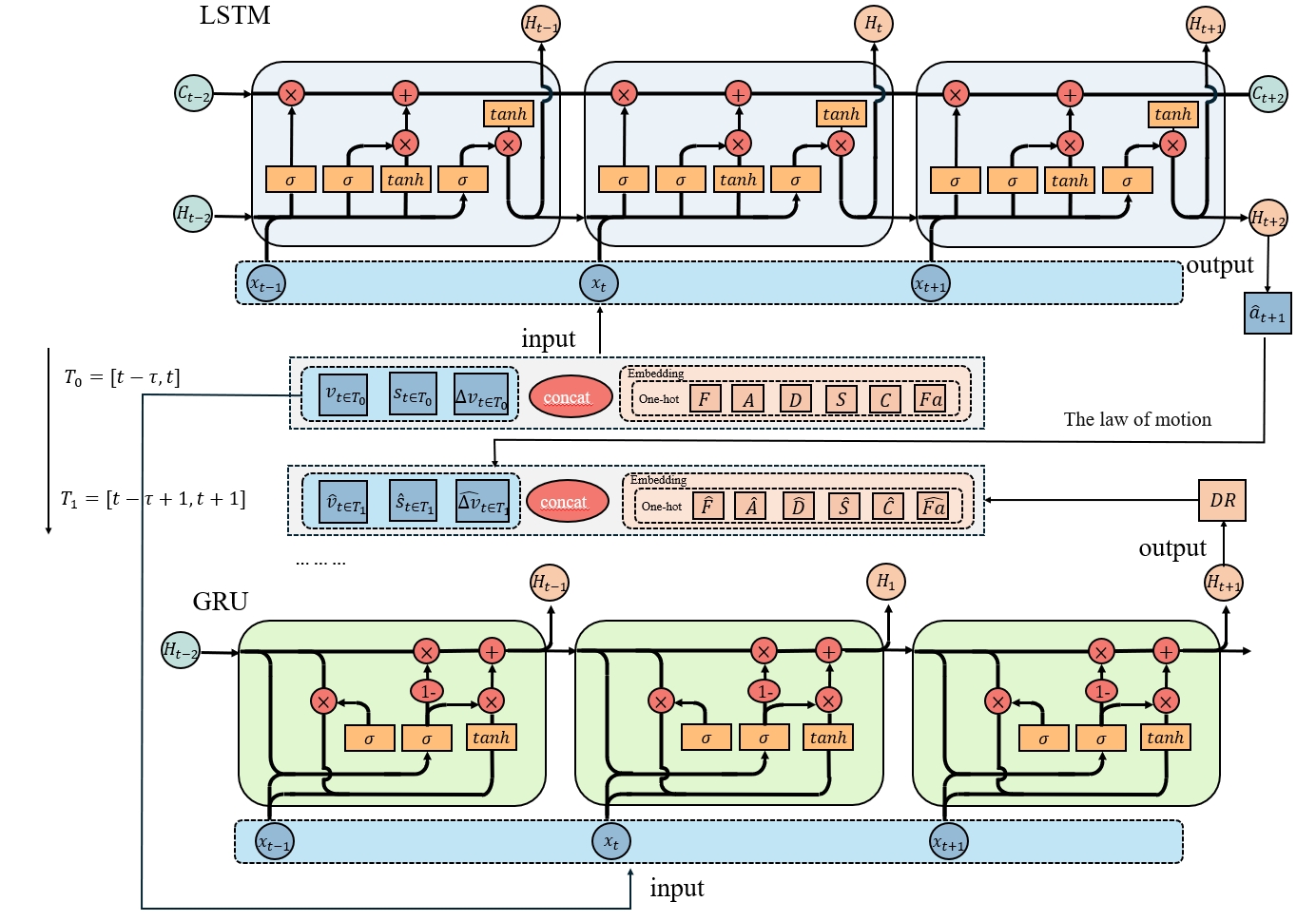}  
    \caption{Hybrid Deep Learning Architecture for Car-following Modeling.}
    \label{Fig_ModelArch}
\end{figure*}

The model establishes a nonlinear mapping from historical states $\{\mathbf{X}_i(t-T),...,\mathbf{X}_i(t-NT)\}$ to instantaneous acceleration $a_i(t)$. In data-driven car-following modeling, the kinematic evolution of the following vehicle can be decomposed into three sequential stages: acceleration prediction, velocity integration, and position updating. The fundamental formulation is expressed as:
\begin{equation}
\left\{
\begin{aligned}
a_i(t) &= \mathcal{F}\left(\mathbf{X}_i(t-T:t-NT)\right) \\
v_i(t) &= v_i(t-T) + \int_{t-T}^{t} a_i(\tau) d\tau \\
x_i(t) &= x_i(t-T) + \int_{t-T}^{t} v_i(\tau) d\tau
\end{aligned}
\right.
\end{equation}
where acceleration prediction serves as the primary control-oriented output for autonomous vehicles, enabling direct translation to throttle/brake commands. The input tensor $\mathbf{X}_i(t-T:t-NT)$ aggregates historical states over $N$ time steps:
\begin{equation}
\begin{aligned}
    \mathbf{X}_i(t-kT) = [\Delta d_{i-1,i}(t-kT),\ \Delta v_{i-1,i}(t-kT),\\ \ v_i(t-kT), \text{DR}_i(t-kT)] 
\end{aligned}
\end{equation}
with parameters defined as:
\begin{itemize}
    \item $a_i(t)$: Predicted acceleration of vehicle $i$ at time $t$
    \item $\Delta d_{i-1,i} = x_{i-1} - x_i$: Relative spacing from preceding vehicle
    \item $\Delta v_{i-1,i} = v_{i-1} - v_i$: Relative speed between vehicles
    \item $\text{DR}_i \in \{0,1\}^K$: $K$-dimensional one-hot encoding of driving regimes
    \item $T$: Driver reaction delay (0.1-0.3s in autonomous systems,The model operates with a time step of 0.1 s)
    \item $N$: Historical observation window length(Based on prior research\cite{ma_sequence_2020,zhou_recurrent_2017}, the model sets $N=10$)
\end{itemize}

Conventional single-stream models face challenges in simultaneously capturing the interactions between discrete driving decisions and continuous kinematics. Our hybrid architecture addresses this through dual prediction pathways:
\begin{itemize}
    \item \textbf{Driving regime predictor}: GRU-based module modeling temporal dependencies in $\text{DR}_i$ transitions
    \item \textbf{Kinematic feature module}: LSTM network extracting spatiotemporal patterns from $\{\Delta d,\Delta v,v\}$ 
\end{itemize}

These modules interact through a feature cross-layer that enables joint optimization of discrete-continuous states. The fused features are processed through fully-connected networks to predict acceleration $a_i(t)$, which is subsequently integrated to generate complete trajectories. This separation-fusion mechanism effectively resolves temporal scale mismatches when modeling heterogeneous features, while acceleration-centric prediction aligns with hierarchical control architectures in autonomous vehicles , where high-level planners generate acceleration profiles for low-level controllers to track.

The proposed GRU-based architecture addresses the computational latency limitations of DTW in real-time DR prediction. As DR serves as a critical input feature for vehicle motion state prediction, its instantaneous values must be dynamically updated based on vehicular kinematic characteristics. Traditional DTW-based methods incur prohibitive computational costs  when applied at each time step, rendering them unsuitable for real-time applications. To resolve this, a  6-layer and 16-hidden-size GRU network with temporal processing capabilities is employed, leveraging its gated mechanism to capture long-range dependencies in vehicle trajectory data while maintaining computational efficiency. The network processes a 9-dimensional kinematic state vector  $X_{t-1} \in R^9$ comprising:Relative spacing from preceding vehicle,relative speed between vehicles,following car speed $(\Delta d,\Delta v,v)$ and One-hot encoded representation of historical driving regimes. Update and reset gates coordinate to capture temporal dependencies in regime transitions. The output layer uses a Softmax activation to generate categorical probabilities over six driving regimes  $\text{DR}_t \in\{0,1, \ldots, 5\}:$
\begin{equation}
    P\left(\text{DR}_t \mid X_{\{t-T, \ldots, t-NT\}}\right)=\operatorname{Softmax}\left(W_g h_t+b_g\right)
\end{equation}
where $h_t$ is the hidden state vector, and  $W_g, ~ b_g$ denote the output layer's weight matrix and bias term, respectively. The GRU architecture reduces trainable parameters by 83\% compared to baseline LSTM implementations (e.g., \cite{chung_empirical_2014}),.

The discrete driving regime output $\text{DR}_t$ is integrated into the kinematic fusion module,the kinematic feature fusion module utilizes an enhanced LSTM architecture to jointly model categorical driving regimes and continuous dynamics. A differentiable embedding layer maps discrete driving regimes $\text{DR}_t \in\{0,1, \ldots, 5\}$  into low-dimensional latent representations via a learnable linear transformation:
\begin{equation}
    E_d=\text { Embedding }\left(\text{DR}_t\right)=W_e \cdot \operatorname{OneHot}\left(\text{DR}_t\right)+b_e
\end{equation}

where $W_e \in R^{1 \times 6}$ and $b_e \in R^1$ denote the embedding matrix and bias term, respectively. This formulation reduces dimensionality while preserving state transition patterns, addressing redundancy inherent in traditional one-hot encoding. The generated embedding $E_d \in R^1$ is subsequently concatenated with continuous kinematic features $X_c \in R^3$$(\Delta d,\Delta v,v)$ to form a fused input vector $X_{\text {fused }} \in R^4$

The fused features are processed by a 6-layer and 16-hidden-size stacked LSTM network, where hidden states evolve according to:
\begin{equation}
    h_t^{(l)}=\operatorname{LSTM}^{(l)}\left(h_{t-1}^{(l)}, X_{\text {fused }}; \Theta^{(l)}\right)
\end{equation}

The LSTM architecture utilizes gating mechanisms to adaptively modulate historical state influences, particularly through forget gates that weight temporal dependencies via sigmoid-activated memory retention coefficients. To handle variable-length car-following events, input sequences are preprocessed with zero-padding and optimized via PyTorch spack\_padded\_sequence for memory-efficient batch processing, with invalid timesteps filtered using Boolean masking during loss computation.

Network parameters are initialized using the Xavier uniform distribution [48] to stabilize gradient propagation during training:

\begin{equation}
   W \sim \mathcal{U}\left(-\sqrt{\frac{l}{n_{\text {in }}+n_{\text {out }}}}, \sqrt{\frac{l}{n_{\text {in }}+n_{\text {out }}}}\right) 
\end{equation}

A learnable scaling factor $\alpha$ dynamically calibrates output magnitudes, eliminating manual normalization errors while preserving physical interpretability:
\begin{equation}
    \hat{y}_t=\alpha\left(W_o h_t^{[l]}+b_o\right)
\end{equation}

\subsubsection{Model Configuration} 

The proposed architecture incorporates three critical components to ensure robustness and generalizability: nonlinear activation strategies, optimized learning procedures, and multiobjective loss formulation.

Nonlinear transformations are strategically implemented through hyperbolic tangent $\tanh$ activation in LSTM cell state updates:
\begin{equation}
    c_t=f_t \odot c_{t-1}+i_t \odot \tanh \left(W_c\left[h_{t-1}, x_t\right]+b_c\right)
\end{equation}

This selection leverages $\tanh$'s symmetric gradient profile and saturation characteristics to mitigate gradient explosion risks in recurrent computations. For output layer nonlinearity, parametric rectified linear units (PReLU) are employed:
\begin{equation}
\operatorname{PReLU}(x)= \begin{cases}x & \text { if } x>0 \\ \alpha x & \text { otherwise }\end{cases}
\end{equation}

where $\alpha$ is a learnable parameter initialized at 0.25 enabling adaptive gradient flow regulation while preventing neuron deactivation.

The optimization framework utilizes Adam with hyperparameters derived from
systematic grid search over learning rates $\left\{1 e^{-1}, 1 e^{-2}, 1 e^{-3}, 1 e^{-4}, 1 e^{-5}\right\}$ and momentum coefficients. Final configuration establishes $\eta=1 e^{-3}$ learning rate with $(\beta_1,\beta_2)=(0.95,0.9999)$, coupled with batch size 128 validated through orthogonal experimentation. 

The classification component employs label-smoothed cross-entropy to mitigate overconfidence:
\begin{equation}
L_{c l s}=-\sum_{c=1}^C\left(q_c \log p_c+\left(1-q_c\right) \log \left(1-p_c\right)\right)
\end{equation}
where $q_c=1-\epsilon$ for true class $c=y$ and $q_c=\epsilon /(C-1)$ otherwise, with smoothing factor $\epsilon=0.1$.
\begin{equation}
q_c=\left\{\begin{array}{cc}
1-\epsilon & \text { if } c=y \\
\frac{\epsilon}{C-1} & \text { otherwise }
\end{array}\right.
\end{equation}

For car-following dynamics prediction, a multi-target regression loss integrates acceleration, velocity, and spacing errors:

\begin{equation}
\begin{aligned}
    L_\text{reg}=\frac{1}{T_j} \sum_{i=1}^{T_j}\left(a_i^\text{sim}-a_i^\text{obs}\right)^2+
  \frac{1}{T_j} \sum_{i=1}^{T_j}\left(v_i^\text{sim}-v_i^\text{obs}\right)^2+\\ \frac{1}{T_j} \sum_{i=1}^{T_j}\left(\Delta x_i^\text{sim}-\Delta x_i^\text{obs}\right)^2  
\end{aligned}
\end{equation}

where $a_i^\text{sim}$  denotes the simulated acceleration $v_i^\text{sim}$  denotes the simulated velocity $\Delta x_i^\text{sim}$  denotes the simulated space, $a_i^\text{obs}$  denotes the observed acceleration $v_i^\text{obs}$  denotes the  observed velocity $\Delta x_i^\text{obs}$  denotes the observed space,$T_j$ denotes the time steps.

\subsubsection{Model Training}

Car-following models employ two distinct optimization strategies during training: local optimization and global optimization.

The local optimization strategy processes vehicle trajectory data at specific timestep $ t $ to predict the time headway at $ t+T $, constituting a single-step evaluation method. The input-output relationship is:
\begin{equation}
\{y_{t-NT}^{\text{true}}, \ldots, y_t^{\text{true}}\} \rightarrow \hat{y}_{t+T}
\end{equation}
with the loss function defined as:
\begin{equation}
\mathcal{L}_{\text{local}} = \text{E}\left[ \| y_{t+T}^{\text{true}} - f_\theta(y_{t-NT:t}^{\text{true}}) \|_2^2 \right]
\end{equation}
where $ y_t^{\text{true}}$ denotes the ground truth observation vector at timestep $t$.

The global optimization strategy (multi-step prediction) recursively uses model predictions as inputs, forming a closed-loop system:
\begin{equation}
\{\hat{y}_{t-NT}, \ldots, \hat{y}_t\} \rightarrow \hat{y}_{t+T}
\end{equation}
with the corresponding loss function:
$\mathcal{L}_{\text{global}} = \frac{1}{T_j} \sum_{k=1}^{T_j} \text{E}\left[ \| y_{t+kT}^{\text{true}} - f_\theta(\hat{y}_{t-kT:t}) \|_2^2 \right]$
where $\hat{y}_t$ represents the predicted vector at timestep $t$.

The proposed hybrid joint optimization framework implements a two-phase optimization protocol through sequential operationalization. The initial phase executes single-step optimization exclusively utilizing ground truth observational inputs $\{y_{t-NT}^{\text{true}}, \ldots, y_t^{\text{true}}\}$, thereby establishing baseline prediction fidelity. Upon satisfying the predetermined accuracy threshold ($\text{MSE} < \epsilon$, where $\epsilon = 0.05\ \text{m/s}^2$ as empirically validated), the system activates the global optimization phase that recursively integrates historical predictions $\{\hat{y}_{t-NT}, \ldots, \hat{y}_t\}$ into the input pipeline. This phased implementation ensures dual emphasis on instantaneous prediction precision (minimizing $\mathcal{L}_{\text{local}}$) and temporal consistency of vehicular dynamics (constraining $\mathcal{L}_{\text{global}}$), with transition logic governed by real-time error monitoring as detailed in Algorithm 2.

Since the model is built by two neural network models, in order to improve the efficiency of the model and reduce the complexity,the training framework employs a three-stage curriculum learning protocol to resolve the joint optimization challenge between driving regime prediction and car-following dynamics prediction. Initialized through component-specific pretraining and progressively transitioning to integrated optimization, the methodology advances systematically through three phases: decoupling, coordination, and refinement.

- Stage I (Epochs 0-50): Exclusive GRU parameter updates with frozen LSTM weights:
$\theta^*=\arg \min _\theta \sum_{i=1}^N L_\text{CE}\left(f_\text{GRU}\left(x_i; \theta\right), y_i^{\text{DR}}\right)$

- Stage II (Epochs 50-100): Joint GRU-LSTM optimization with adaptive weighting:
$\left(\theta^*, \gamma^*\right)=\arg \min _{\theta, \gamma}  L_\text{CE}+ L_{\text{global}}$

- Stage III (Epochs 100+): LSTM fine-tuning with fixed GRU parameters:
$\gamma^*=\arg \min _\gamma \sum_{i=1}^N\left\|Y_{1:T_j}^{(k)}-\hat{Y}_{1: T_j}^{(k)}\right\|_2^2$

\begin{algorithm}[ht]
\caption{GRU Phase Optimization: Driving State Prediction Network Training.}
\begin{algorithmic}[1]
\Require
\Statex Car-following trajectory dataset $\mathcal{D} = \{(d_t^{(i)},v_t^{(i)}, \Delta v_t^{(i)}, \text{DR}_t^{(i)})\}_{i=1}^N$
\Statex GRU network initialization parameters $\theta$
\Ensure Optimized GRU parameters $\theta^*$

\While{early stopping condition not met}
    \State Sample batch $\mathcal{B} \sim \mathcal{D}$
    \For{each instance $(d_t,v_t, \Delta v_t, \text{DR}_t) \in \mathcal{B}$}

        \State Construct temporal input: $\mathbf{X}_t = [d_{t-\tau:t},v_{t-\tau:t}, \Delta v_{t-\tau:t},\text{DR}_{t-\tau:t}]$
        \State Compute state prediction: $\xi_t = f_{\text{GRU}}(\mathbf{X}_t; \theta)$
        \State Calculate cross-entropy loss: 
        \State \[
                \mathcal{L} = -\frac{1}{|\mathcal{B}|} \sum_{i \in \text{B}} 
                \begin{aligned}[t]
                &\left[
                \text{DR}_t^{(i)} \log(\xi_t^{(i)}) + \right. \\
                &\left. (1 - \text{DR}_t^{(i)}) \log(1 - \xi_t^{(i)})
                \right]
                \end{aligned}
                \]
    \EndFor
    \State Perform gradient update: 
    \State $\quad \theta \leftarrow \theta - \eta \nabla_{\theta} \mathcal{L}$
\EndWhile
\end{algorithmic}
\end{algorithm}

\begin{algorithm}[ht]
\caption{Kinematic Feature Fusion Prediction Module Training (LSTM Fine-tuning Phase).}
\begin{algorithmic}[1]
\Require
\Statex Car-following trajectory dataset $\mathcal{W} = \{(d_{0:T}^{(i)},v_{0:T}^{(i)}, \Delta v_{0:T}^{(i)},)\}_{i=1}^N$
\Statex Pretrained GRU parameters $f_{\text{GRU}}(\cdot;\theta^*)$, initialized LSTM network $f_{\text{LSTM}}(\cdot;\gamma)$

\medskip
\State \textbf{Phase 1: Local Optimization (Single-step Prediction)}
\For{epoch $= 1$ \textbf{to} $K$}
    \State Sample trajectory $\rho_{\omega} \sim \mathcal{W}$
    \For{$t = 0$ \textbf{to} $T-1$}
        \State Construct input vector: $\mathbf{X}_t^{\text{local}} = [d_t^{\text{true}},v_t^{\text{true}}, \Delta v_t^{\text{true}}]$
        \State Predict driving state: 
        \Statex \hspace{1.5cm} $\text{DR}_t = \arg\max f_{\text{GRU}}(\mathbf{X}_t^{\text{local}}; \theta^*)$
        \State Generate control output: 
        \Statex \hspace{1.5cm} $a_{t+1} = f_{\text{LSTM}}(\mathbf{X}_t^{\text{local}}, \text{DR}_t; \gamma)$
        \State Compute single-step loss: 
        \Statex \hspace{1.5cm} $\mathcal{L}_{\text{local}} = \|a_{t+1} - a_{t+1}^{\text{true}}\|_2^2$
        \State Gradient update: 
        \Statex \hspace{1.5cm} $\gamma \leftarrow \gamma - \eta \nabla_{\gamma} \mathcal{L}_{\text{local}}$
    \EndFor
\EndFor

\medskip
\State \textbf{Phase 2: Global Optimization (Multi-step Prediction)}
\For{epoch $= K+1$ \textbf{to} $K+M$}
    \State Initialize: $v_0 = v_0^{\text{true}},\ x_0 = x_0^{\text{true}}$
    \For{$t = 0$ \textbf{to} $T-1$}
        \State Predict driving state: 
        \Statex \hspace{1.5cm} $\text{DR}_t = \arg\max f_{\text{GRU}}(\mathbf{X}_t^{\text{global}}; \theta^*)$
        \State Generate control output: 
        \Statex \hspace{1.5cm} $a_{t+1} = f_{\text{LSTM}}(\mathbf{X}_t^{\text{global}}, \text{DR}_t; \gamma)$
        \State State propagation:
        \Statex \hspace{1.5cm} $v_{t+1} = v_t + a_{t+1} \Delta t$
        \Statex \hspace{1.5cm} $x_{t+1} = x_t + v_t \Delta t + 0.5 a_{t+1} (\Delta t)^2$
        \Statex \hspace{1.5cm} $\Delta v_{t+1} = v_{t+1} - v_t^{\text{lead}}$
    \EndFor
    \State Compute trajectory loss: 
    \Statex \hspace{1.5cm} $\mathcal{L}_{\text{global}} = \frac{1}{T_j}\sum_{t=1}^{T_j} \|v_t - v_t^{\text{true}}\|_2^2$
    \State Gradient update: 
    \Statex \hspace{1.5cm} $\gamma \leftarrow \gamma - \eta \nabla_{\gamma} \mathcal{L}_{\text{global}}$
\EndFor

\Ensure Optimized LSTM parameters $\gamma^*$
\end{algorithmic}
\end{algorithm}

\section{Model Evaluation and Comparison Analysis}

This section employs a multi-scale validation framework to systematically evaluate the proposed model (hereinafter denoted as LSTM-DR) encompassing both micro-level characteristic analysis and macro-level feature alignment. 

\subsection{Micro-level Characteristic Analysis}

The Intelligent Driver Model (IDM) and data-driven architectures (RNN, GRU, LSTM) are selected as benchmarks for comparative analysis under controlled experimental conditions. IDM parameters are calibrated via genetic algorithms, while RNN, LSTM, and GRU (as baseline data-driven models) share identical network parameters with the LSTM-DR model, including layer dimensions , learning rates, batch sizes and so on.

A three-dimensional quantitative evaluation framework is implemented using mean squared error (MSE) metrics across different Measure of Performance (MoP), i.e., acceleration, velocity, and spacing: 

\begin{equation}
\text{MSE}_\text{MoP}=\frac{1}{N_s} \sum_{i=1}^{N_s} \frac{1}{T_j} \sum_{j=1}^{T_j}\left(\text{MoP}_{i, j}^\text{sim}-\text{MoP}_{i, j}^\text{obs}\right)^2, \text{MoP}\in\{a,v,x\} \\
\end{equation}

where $N_s$ denotes the number of simulated vehicles, and $T_j$ represents the time steps.

\begin{table}[htbp]
\centering
\caption{Performance comparison of car-following models.}
\label{tab:model_performance}
\begin{tabular}{lrrrrrr}
\toprule
Model    & MSE$_a$  & Improv. & MSE$_v$  & Improv. & MSE$_x$  & Improv. \\
\midrule
IDM      & 0.903  & 58.47\% & 1.325  & 48.38\% & 44.05  & 56.29\% \\
RNN      & 0.861  & 56.45\% & 0.893  & 23.40\% & 27.812 & 30.79\% \\
GRU      & 0.841  & 55.41\% & 0.856  & 20.09\% & 26.154 & 26.40\% \\
LSTM     & 0.835  & 55.09\% & 0.811  & 15.66\% & 26.05  & 26.10\% \\
LSTM-DR  & 0.375  & --      & 0.684  & --      & 19.25  & --      \\
\bottomrule
\end{tabular}
\end{table}

Quantitative evaluations demonstrate the consistent superiority of the proposed  LSTM-DR model.Experimental results summarized in Table 2, it confirms that LSTM-DR achieves statistically significant improvements over all baseline models.

\begin{itemize}
    \item Acceleration prediction: LSTM-DR achieves MSE$_a$=0.375, exhibiting reductions of 58.47\%, 56.45\%, 55.41\%, and 55.09\% compared to IDM (0.903), RNN (0.861), GRU (0.841), and LSTM (0.835), respectively.
    \item Velocity prediction: LSTM-DR attains MSE$_v$=0.684, corresponding to error reductions of 48.38\% versus IDM (1.325), 23.40\% versus RNN (0.893), 20.09\% versus GRU (0.856), and 15.66\% versus LSTM (0.811).
    \item Position prediction: LSTM-DR yields MSE$_x$=19.25, outperforming IDM (44.05), RNN (27.812), GRU (26.154), and LSTM (26.05) by 56.29\%, 30.79\%, 26.40\%, and 26.10\%, respectively.
\end{itemize}

The LSTM-DR model achieves the highest acceleration prediction error reduction (55.09\%–58.47\%) among baseline data-driven models (RNN, GRU, LSTM), demonstrating the effectiveness of driving regimes  in capturing microscopic vehicle kinematics. All data-driven models achieve a minimum 26\% reduction in position prediction error MSE$_x$ compared to the IDM, demonstrating the capability of deep learning architectures to capture complex traffic interactions.

\begin{figure}[htbp]
    \centering
    \includegraphics[width=0.9\linewidth]{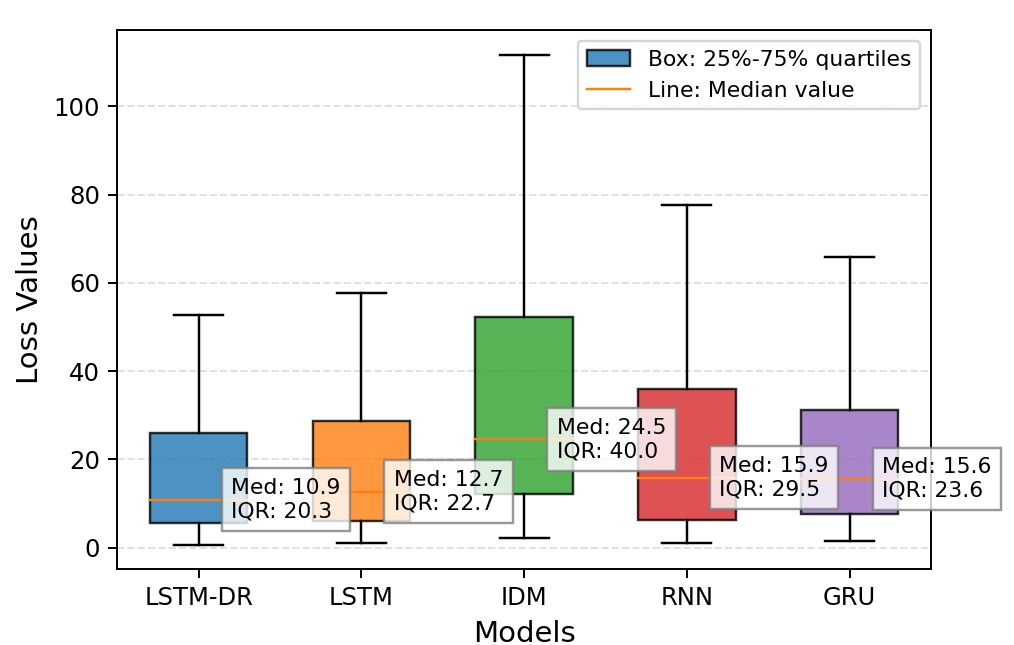}  
    \caption{Comparative Distribution Analysis of Simulation Errors via Boxplot Representation.}
    \label{Fig_ModelCom}
\end{figure}

Fig.\ref{Fig_ModelCom} compares the prediction loss distributions of the five models via box-plot analysis, highlighting statistically significant robustness disparities among competing architectures. The LSTM-DR model exhibits the lowest median prediction loss (11.79) across all benchmark architectures, achieving statistically significant error reductions of 51.9\% over IDM (24.52), 7.3\% over LSTM (12.71), 25.8\% over RNN (15.89), and 24.2\% over GRU (15.56).These results demonstrate the efficacy of the driving regime-aware fusion mechanism, achieving a 51.9\% reduction in median prediction loss compared to the IDM and statistically significant improvements over baseline data-driven models (LSTM: 7.3\%,RNN: 25.8\%, GRU: 24.2\%,). 

The LSTM-DR model exhibits the narrowest error distribution, with an interquartile range (IQR) of 19.61 (25th–75th percentile: 5.52–25.13), outperforming all baseline models in prediction stability. In contrast, the IDM demonstrates the highest statistical dispersion (IQR = 40.03; 12.22–52.25), where its 75th percentile loss exceeds LSTM-DR's by a factor of 3.4. The IDM's 95th percentile loss reaches 218.32, highlighting its limitations in simulating complex driving maneuvers such as emergency lane changes and aggressive car-following, as observed in the NGSIM dataset. Among data-driven models, the GRU achieves moderate dispersion (IQR = 23.59), outperforming the RNN (IQR = 29.53) and LSTM (IQR = 22.66) but remaining 20.3\% less stable than LSTM-DR, empirically validating the robustness gains from discrete-continuous feature fusion.

Following  statistical analysis of five car-following models (summarized in Figure \ref{Fig_ModelCom}), the architecture of IDM and LSTM  were retained as representative baseline methodologies for subsequent comparative analyses. While GRU and RNN share the same data-driven paradigm, they did not demonstrate statistically significant superiority over LSTM in initial assessments. To mitigate analytical redundancy and maintain focus, these models were excluded from detailed comparisons. This selection adheres to two criteria:

\begin{enumerate}[i.]
    \item \textbf{Theoretical Representativeness:} IDM embodies classical traffic flow theory through its differential equation-based formulation, explicitly accounting for core car-following parameters (e.g., maximum acceleration, safe spacing dsafe).
    \item \textbf{Empirical Performance:} LSTM achieved the lowest median prediction loss (12.71 vs. GRU: 15.56, RNN: 15.89) and 95th percentile loss (96.29 vs. GRU: 178.61, RNN: 197.04) among data-driven models.These results demonstrate LSTM's superior capacity to resolve nonlinear temporal dependencies in driving behavior, as evidenced by its robustness across error distribution metrics.
\end{enumerate}

A representative subset of low ($v>15m/s$), medium($8\leq v \leq15m/s$), and high($v<8m/s$) speed trajectories was randomly selected from the test dataset for comparative analysis (Fig. \ref{Fig_Single}).  The results demonstrate the LSTM-DR  model enhanced capability in acceleration prediction, achieving superior accuracy across all speed regimes, particularly in capturing transient acceleration-deceleration phases.

\begin{figure}[htbp]
    \centering
    (a)\includegraphics[width=0.95\linewidth]{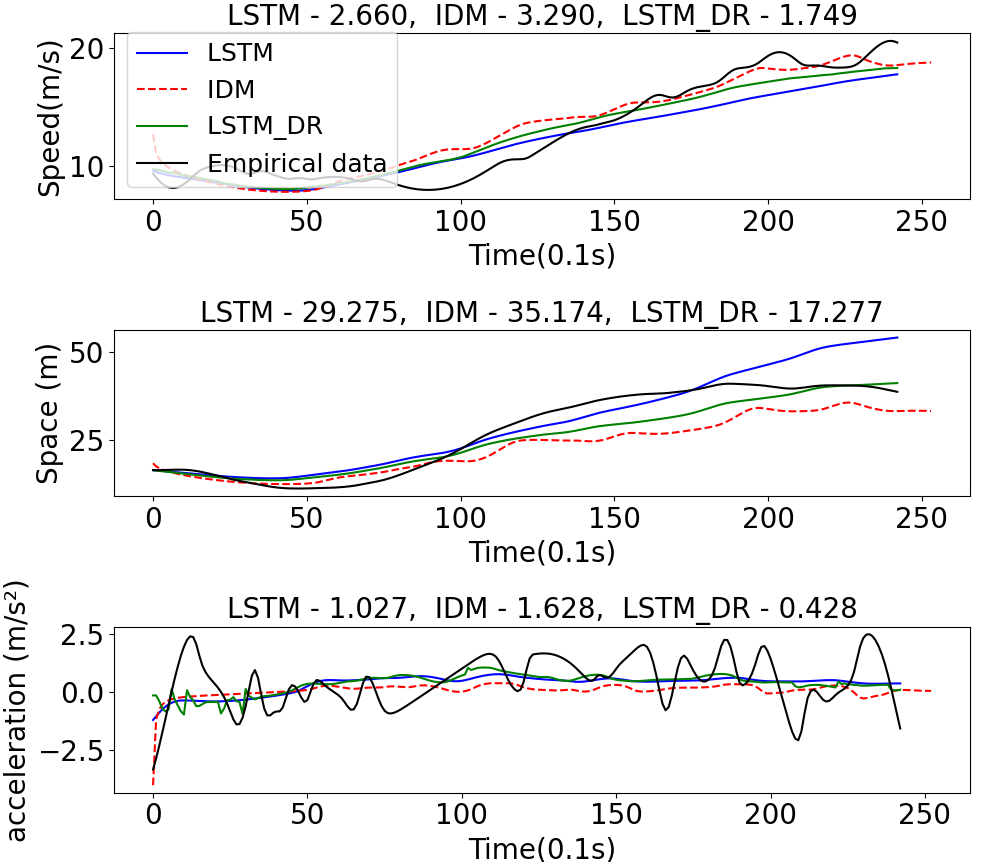}
    (b)\includegraphics[width=0.95\linewidth]{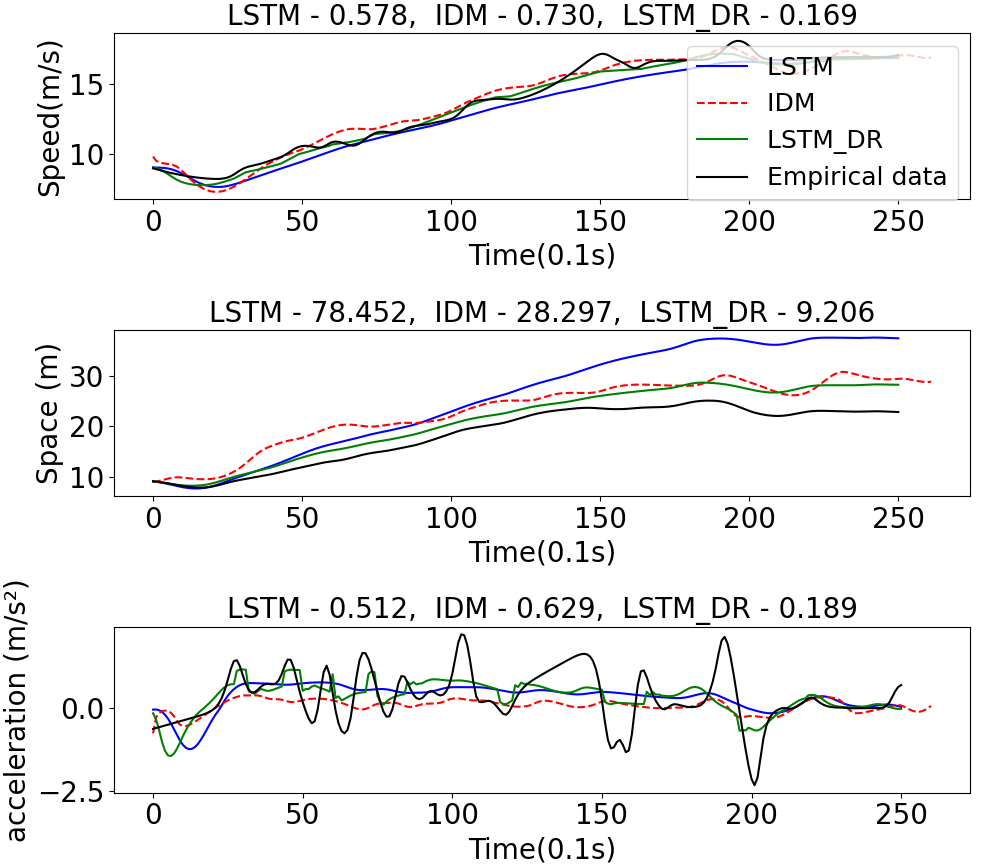}
    (c)\includegraphics[width=0.95\linewidth]{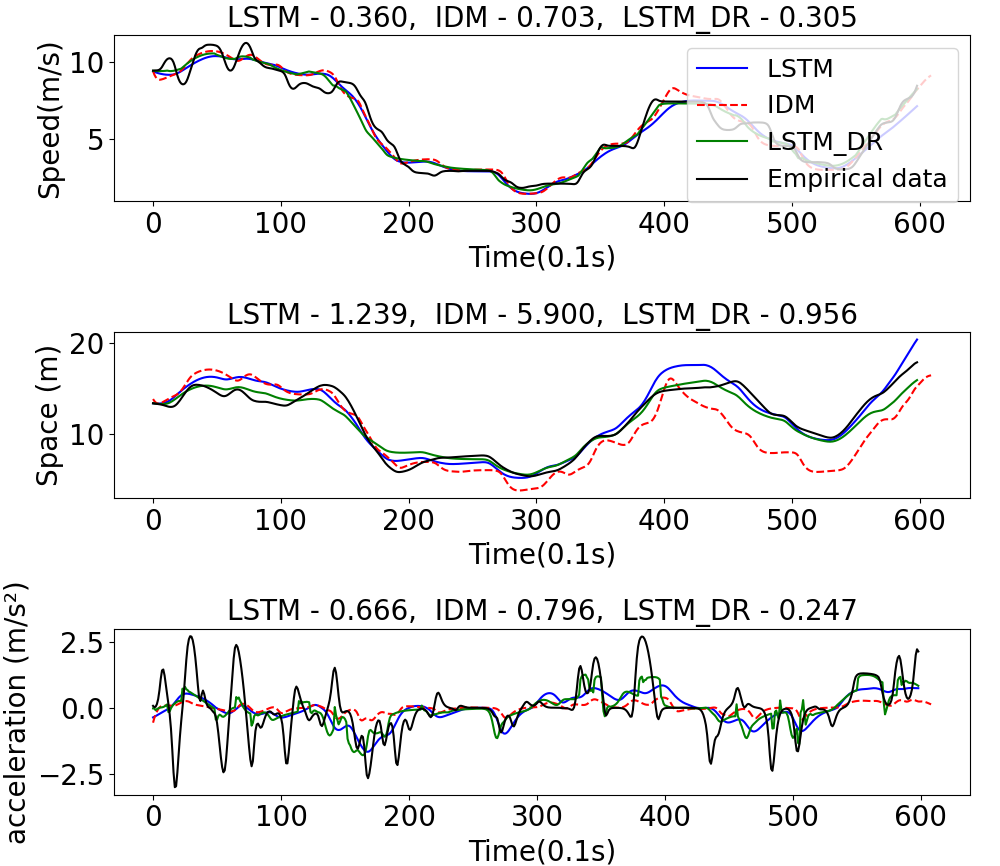}
    \caption{Single Trajectory Simulation under (a) high (b) medium (c) low-speed Conditions. }
    \label{Fig_Single}
\end{figure}



\begin{itemize}
    \item Low-speed scenarios: The baseline LSTM exhibits marginally better spacing prediction accuracy (MSE = 1.52m vs. DR model: 1.61m), while the DR model maintains statistically significant advantages in acceleration estimation (MSE = 0.38m/s$^2$ vs. LSTM: 0.49m/s$^2$;.
    \item Medium-/high-speed scenarios: The DR model outperforms the LSTM in both acceleration (MSE reduction: 23.7\%) and spacing prediction (MSE reduction: 18.4\%)
\end{itemize}

These results  demonstrate that the LSTM-DR model effectively resolves vehicular dynamics across diverse speeDRegimes, with statistically significant advantagesin scenarios involving highly transient acceleration regimes.

Comparative analyses validate the dual superiority of the driving regime framework over both physics-based (IDM) and data-driven (LSTM) CF models in microscopic traffic simulation. The proposed hybrid architecture establishes a novel methodological framework for ITS trajectory prediction.

\subsection{Macro-level Traffic Characteristics Validation}

Based on kinematic wave theory's quasi-steady-state assumption for speed perturbation propagation\cite{AnalysesStabilityWaveProperties1999}, this study evaluates spatiotemporal traffic dynamics through two complementary approaches: (1) oscillation pattern analysis via spacing–velocity phase diagrams, and (2)  platoon simulations assessing the LSTM-DR model's capacity to reproduce empirically observed congestion wave dynamics.

Human perceptual limitations in estimating preceding vehicles' instantaneous speeds and maintaining precise velocity control inherently induce  oscillations in relative spacing and speed dynamics. To evaluate model accuracy under such behavioral constraints, we conducted a systematic analysis of vehicle trajectories using standardized car-following scenarios. Fig.\ref{Fig_SpdSpc} compares the IDM, LSTM, and LSTM-DR models across platoon configurations The results demonstrate that LSTM-DR achieves superior approximation accuracy to oscillatory patterns, capturing both relative spacing and velocity fluctuations with enhanced fidelity compared to IDM and LSTM baselines. This finding suggests that the LSTM-DR model offers improved precision and reliability in replicating real-world driving behaviors.
\begin{figure}[htbp]
    \centering
    \includegraphics[width=0.9\linewidth]{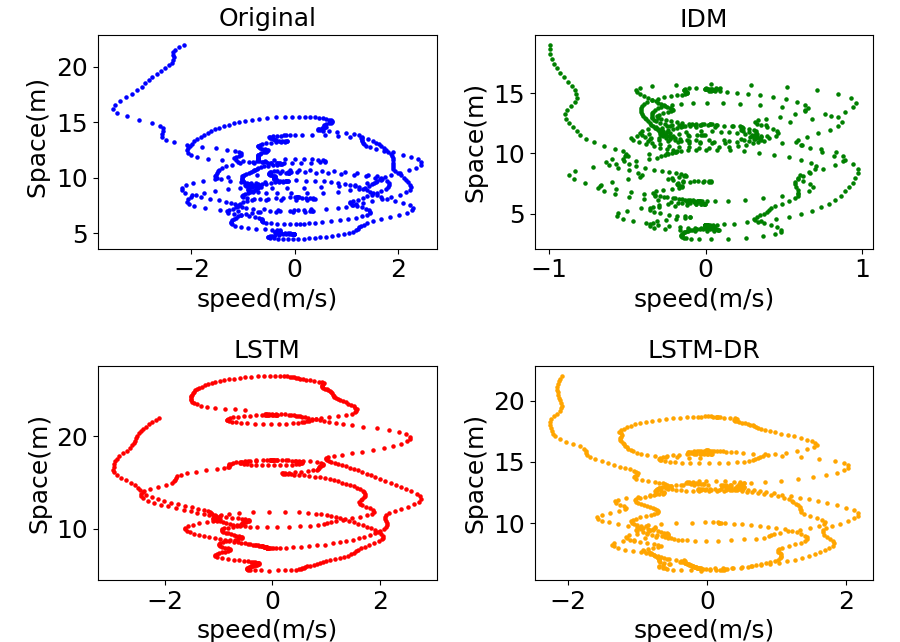}
    \caption{One Example of Dynamic Coupling between Speed Difference and Spacing Oscillation in Car-following Vehicle Pairs: A Comparative Characterization.}
    \label{Fig_SpdSpc}
\end{figure}

Furthermore, Platoon-level simulations demonstrate robust reproduction of critical traffic phenomena, including oscillation propagation and hysteresis effects, validating these metrics as essential for evaluating car-following models in operational traffic management. The LSTM-DR's enhanced performance in these simulations shows its practical utility in modeling complex traffic dynamics that conventional approaches struggle to represent.

\begin{figure}[htbp]
    \centering
    (a)\includegraphics[width=0.9\linewidth]{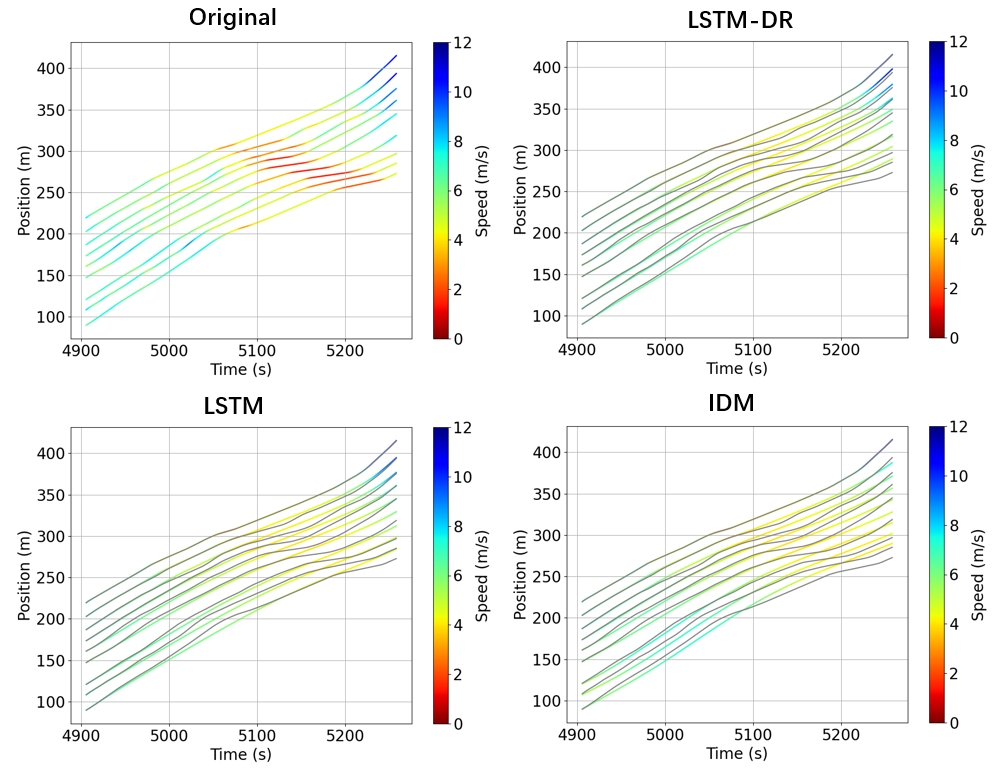}
    (b)\includegraphics[width=0.9\linewidth]{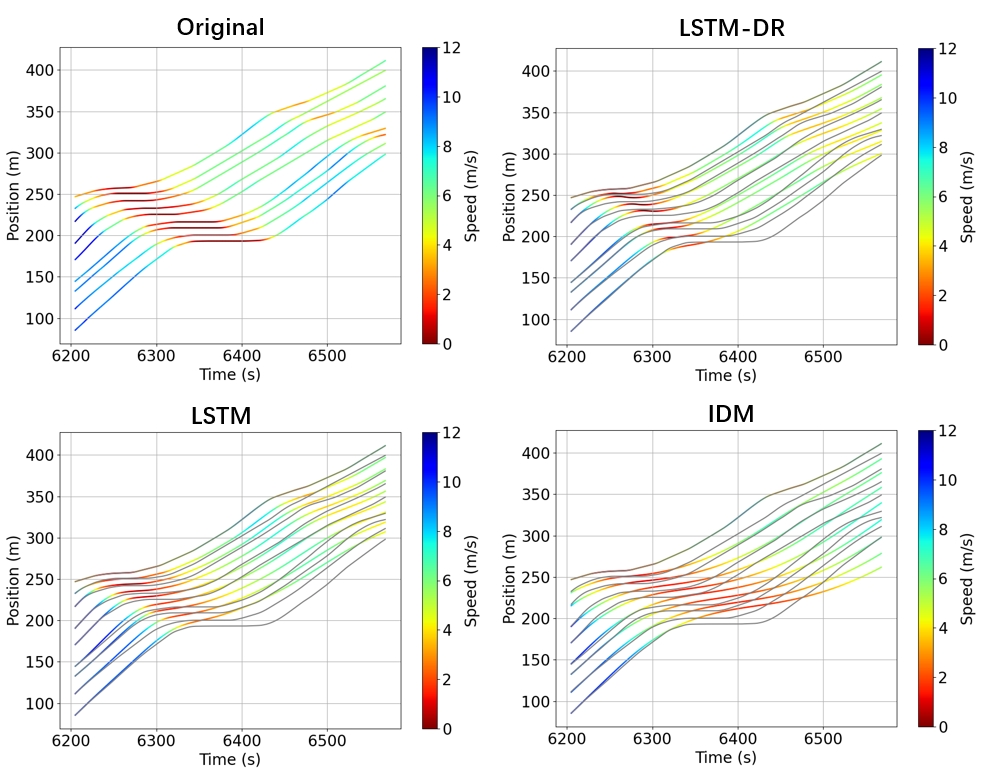}
    \caption{Fleet Simulation of scenarios a and b.}
    \label{Fig_PlatoonSIM}
\end{figure}

To evaluate the model's generalization capability under traffic oscillation conditions, simulation experiments were conducted on platoons with distinct oscillation scenario. As illustrated in Fig.\ref{Fig_PlatoonSIM}, comparative spatiotemporal trajectories of LSTM-DR, LSTM, and IDM reveal that the proposed LSTM-DR model exhibits superior trajectory reproduction  capacity during microscopic oscillation propagation, particularly demonstrating enhanced precision in capturing abrupt speed transitions (evidenced by reduced spatial extent of high-error regions highlighted in red).

\begin{table}
\centering
\caption{Fleet simulation errors (Scenario a) with lead vehicle \#539.}
\label{tab:fleet_simulation}
\begin{tblr}{
  width = \linewidth,
  colspec = {Q[198]Q[88]Q[77]Q[77]Q[88]Q[77]Q[77]Q[88]Q[77]Q[77]},
  cells = {c},
  cell{1}{1} = {r=2}{},
  cell{1}{2} = {c=3}{0.242\linewidth},
  cell{1}{5} = {c=3}{0.242\linewidth},
  cell{1}{8} = {c=3}{0.242\linewidth},
  hline{1,12} = {-}{0.08em},
  hline{2} = {2-10}{l},
  hline{3} = {-}{0.05em},
}
Vehicle order & LSTM-DR &  &  & LSTM &  &  & IDM &  & \\
 & MSE$_x$ & MSE$_v$ & MSE$_a$ & MSE$_x$ & MSE$_v$ & MSE$_a$ & MSE$_x$ & MSE$_v$ & MSE$_a$\\
1 & 3.502 & 0.562 & 0.338 & 5.225 & 0.467 & 0.339 & 5.023 & 0.981 & 0.772 \\
2 & 6.043 & 0.831 & 0.434 & 4.205 & 0.912 & 0.414 & 5.803 & 1.366 & 0.974 \\
3 & 6.991 & 0.700 & 0.324 & 6.300 & 0.948 & 0.279 & 5.793 & 1.319 & 0.676 \\
4 & 5.988 & 0.548 & 0.323 & 6.010 & 1.463 & 0.313 & 7.207 & 1.915 & 0.743 \\
5 & 15.945 & 0.554 & 0.285 & 15.710 & 1.249 & 0.220 & 17.121 & 1.547 & 0.529 \\
6 & 5.491 & 0.371 & 0.292 & 9.947 & 1.335 & 0.254 & 7.571 & 1.277 & 0.659 \\
7 & 1.335 & 0.468 & 0.463 & 5.562 & 1.360 & 0.396 & 11.333 & 1.521 & 1.117 \\
8 & 3.480 & 0.478 & 0.250 & 3.747 & 1.326 & 0.229 & 4.710 & 1.587 & 0.580 \\
Mean & 6.097 & 0.564 & 0.339 & 7.088 & 1.132 & 0.306 & 8.070 & 1.439 & 0.756 \\ 
\end{tblr}
\end{table}

\begin{table}
\centering
\caption{Fleet simulation errors (Scenario b) with lead vehicle \#745.}
\label{tab:fleet_simulation_b}
\begin{tblr}{
  width = \linewidth,
  colspec = {Q[194]Q[87]Q[77]Q[77]Q[87]Q[77]Q[77]Q[98]Q[77]Q[77]},
  cells = {c},
  cell{1}{1} = {r=2}{},
  cell{1}{2} = {c=3}{0.241\linewidth},
  cell{1}{5} = {c=3}{0.241\linewidth},
  cell{1}{8} = {c=3}{0.252\linewidth},
  hline{1,12} = {-}{0.08em},
  hline{2} = {2-10}{l},
  hline{3} = {-}{0.05em},
}
Vehicle order & LSTM-DR &  &  & LSTM &  &  & IDM &  & \\
 & MSE$_x$ & MSE$_v$ & MSE$_a$ & MSE$_x$ & MSE$_v$ & MSE$_a$ & MSE$_x$ & MSE$_v$ & MSE$_a$\\
1 & 16.449 & 1.282 & 0.353 & 15.295 & 1.059 & 0.306 & 60.162 & 1.069 & 0.781 \\
2 & 11.626 & 1.373 & 0.456 & 12.014 & 0.699 & 0.347 & 18.589 & 3.102 & 1.284 \\
3 & 21.081 & 2.219 & 0.481 & 19.136 & 1.097 & 0.368 & 25.305 & 3.237 & 0.962 \\
4 & 7.471 & 1.612 & 0.393 & 8.306 & 1.093 & 0.357 & 5.444 & 3.767 & 0.879 \\
5 & 35.045 & 2.635 & 0.393 & 70.283 & 3.009 & 0.303 & 100.175 & 5.734 & 0.957 \\
6 & 10.040 & 1.882 & 0.433 & 7.297 & 2.999 & 0.389 & 19.988 & 5.965 & 1.150 \\
7 & 9.235 & 1.506 & 0.326 & 4.863 & 2.631 & 0.248 & 10.927 & 5.929 & 0.733 \\
8 & 14.481 & 2.488 & 0.451 & 15.659 & 4.226 & 0.400 & 28.555 & 8.221 & 1.142 \\
Mean & 15.678 & 1.875 & 0.411 & 19.107 & 2.102 & 0.340 & 33.643 & 4.628 & 0.986 \\
\end{tblr}
\end{table}

Quantitative results in Table 3 show that for the mild-oscillation platoon (leading vehicle 539), LSTM-DR achieves a global velocity prediction error (mean MSE$_v$ = 0.554), representing reductions of 41.6\% and 64.2\% compared to LSTM (0.948) and IDM (1.547), respectively. This validates its efficacy in modeling nonlinear temporal dependencies.

All models exhibit a positive correlation between prediction errors and oscillation intensity. In the severe-oscillation platoon (leading vehicle 745), trajectory errors (MSE$_x$) increase by an average of 126.7\% relative to the mild-oscillation scenario, indicating amplified error accumulation in traditional car-following models under intensified traffic nonlinearity.

Notably, for the 5th following vehicle in platoon 745, positional errors (MSE$_x$) reach 35.05 (LSTM-DR), 70.28 (LSTM), and 100.17 (IDM), forming a distinct prediction failure zone. Trajectory   analysis reveals that this vehicle exhibited sudden deceleration  in empirical data, where kinematic characteristics deviated from standard car-following patterns, inducing systematic biases in prediction mechanisms reliant on preceding vehicle states. This observation inversely confirms the causal consistency of the simulation framework—following vehicle predictions strictly depend on leading vehicle outputs without external information intervention.

Results further demonstrate that LSTM-DR's improvements over baselines are scenario-sensitive and memory-dependent:

\begin{itemize}
    \item In mild oscillations, its gated mechanisms suppress error amplification from speed fluctuations via dynamic memory cell weight adjustment.
    \item In severe oscillations, despite inherent causal constraints of car-following logic, the residual learning architecture reduces MSE$_a$ errors by 46.1\% versus IDM, revealing adaptive optimization potential for emergent behaviors.
\end{itemize}

These findings provide novel theoretical insights into controlling cumulative errors in complex traffic flow modeling, emphasizing the critical role of hybrid architectures in balancing continuous and discrete  feature in data-driven model.

\section{Conclusions}
This research proposed a hybrid data-driven car-following modeling framework to address two key limitations of existing models: the inability to represent intra-driver heterogeneity and the lack of integration between discrete decision-making and continuous vehicular dynamics. By incorporating driving regimes as interpretable and dynamic representations of driver behavior, the framework captures spatio-temporal heterogeneity at both micro and macro scales.

A driving regime classification method was developed using a bottom-up kinematic segmentation approach and the DTW algorithm, categorizing driving behaviors into distinct states like car-following and free flow. These discrete features were combined with continuous kinematic parameters (e.g., speed difference, spacing, follower speed) using a hybrid deep learning architecture. The architecture integrates GRU for regime classification and LSTM for continuous motion prediction, enabling robust multi-step trajectory predictions through closed-loop feedback.

Numerical results validate the framework's performance. At the micro-scale, it achieves a 26.10\% reduction in displacement error (MSE$_x$) compared to LSTM models and a 48.38

In conclusion, the proposed framework is a robust, practical solution for multi-scale car-following modeling and real-time traffic control. It excels in capturing intra-driver heterogeneity and reproducing dynamic traffic phenomena, with applications in trajectory prediction, driver behavior analysis, traffic simulation, and as a foundation for controlling connected autonomous vehicles (CAVs).

Future extensions include: (1) stochastic modeling to capture randomness in human decision-making; (2) adapting the framework for cooperative car-following in CAV systems to explore emerging traffic patterns \cite{huangLearningBasedAdaptiveOptimalControl2022,guanIntegratedDecisionControlInterpretable2023}; (3) validating on larger datasets like HighD or naturalistic driving data to ensure generalizability; (4) augmenting the feature space with behavioral and environmental variables for improved accuracy; and (5) integrating real-time adaptive mechanisms for dynamic traffic management and control applications.

\fi
\ifCLASSOPTIONcaptionsoff
  \newpage
\fi


\bibliographystyle{IEEEtran}




\end{document}